\documentclass[runningheads]{llncs}
\usepackage{graphicx}

\usepackage{tikz}
\usepackage{comment}
\usepackage{amsmath,amssymb} 
\usepackage{bbm}
\usepackage{multirow}
\usepackage{tabu}
\usepackage{color}
\usepackage{arydshln}
\usepackage{mathtools}
\makeatletter
\newcommand{\printfnsymbol}[1]{%
\textsuperscript{\@fnsymbol{#1}}%
}

\begin{document}
\pagestyle{headings}
\mainmatter
\def\ECCVSubNumber{754}  

\title{Part-aware Prototype Network for Few-shot Semantic Segmentation} 

\titlerunning{Part-aware Prototype Network for Few-shot Semantic Segmentation} 
\author{Yongfei Liu\textsuperscript{\rm 1,3,4}\thanks{Both authors contributed equally to the work. This work was supported by Shanghai NSF Grant (No. 18ZR1425100)} \and
Xiangyi Zhang\textsuperscript{\rm 1}\printfnsymbol{1} \and
Songyang Zhang\textsuperscript{\rm 1}\and 
Xuming He\textsuperscript{\rm 1,2}}
\institute{\textsuperscript{\rm 1}School of Information Science and Technology, ShanghaiTech University\\
\textsuperscript{\rm 2}Shanghai Engineering Research Center of Intelligent Vision and Imaging\\
\textsuperscript{\rm 3}Shanghai Institute of Microsystem and Information Technology, Chinese Academy of Sciences \\
\textsuperscript{\rm 4}University of Chinese Academy of Sciences\\
\email{\{liuyf3, zhangxy9, zhangsy1, hexm\}@shanghaitech.edu.cn}}
\authorrunning{Yongfei Liu, Xiangyi Zhang et al.}
\maketitle

\begin{abstract}
  Few-shot semantic segmentation aims to learn to segment new object classes with only a few annotated examples, 
  which has a wide range of real-world applications. Most existing methods either focus on the restrictive setting 
  of one-way few-shot segmentation or suffer from incomplete coverage of object regions. In this paper, we propose 
  a novel few-shot semantic segmentation framework based on the prototype representation. Our key idea is to decompose 
  the holistic class representation into a set of part-aware prototypes, capable of capturing diverse and fine-grained 
  object features. In addition, we propose to leverage unlabeled data to enrich our part-aware prototypes, resulting 
  in better modeling of intra-class variations of semantic objects. We develop a novel graph neural network model to 
  generate and enhance the proposed part-aware prototypes based on labeled and unlabeled images. 
  Extensive experimental evaluations on two benchmarks show that our method outperforms the prior art with a sizable margin.
  \footnote{Code is avilable at: https://github.com/Xiangyi1996/PPNet-PyTorch}

\end{abstract}

\section{Introduction}
Semantic segmentation is a core task in modern computer vision with many potential applications ranging from autonomous navigation~\cite{Brabandere2017Semantic} to medical image understanding~\cite{chung2017learning}. A particular challenge in deploying segmentation algorithms in real-world applications is to adapt to novel object classes efficiently in dynamic environments. Despite the remarkable success achieved by deep convolutional networks in semantic segmentation~\cite{long2015fully,badrinarayanan2017segnet,chen2017rethinking,zhao2017pyramid}, a notorious disadvantage of those supervised approaches is that they typically require thousands of pixel-wise labeled images, which are very costly to obtain. While much effort has been made to alleviate such burden on data annotation, such as weak supervision~\cite{kipf2017semi}, most of them still rely on collecting large-sized datasets. 

A promising strategy, inspired by human visual recognition~\cite{russakovsky2015imagenet}, is to enable the algorithm to learn to segment a new object class with only a few annotated examples. Such a learning task, termed as \textit{few-shot semantic segmentation}, has attracted much attention recently~\cite{rakelly2018conditional,boots2017one,wang2019panet,rakelly2018few}.
Most of those initial attempts adopt the metric-based meta-learning framework~\cite{vinyals2016matching}, in which they first match learned features from support and query images, and then decode the matching scores into final segmentation. 

However, the existing matching-based methods often suffer from several drawbacks due to the challenging nature of semantic segmentation. 
First, some prior works~\cite{zhang2018sg,zhang2019canet,zhang2019pyramid} solely focus on the task of one-way few-shot segmentation. Their approaches employ dense pair-wise feature matching and specific decoding networks to generate segmentation, and hence it is non-trivial or computationally expensive to generalize to the multi-way setting. Second, other prototype-based methods~\cite{dong2018few,siam2019adaptive,wang2019panet} typically use a holistic representation for each semantic class, which is difficult to cope with diverse appearance in objects with different parts, poses or subcategories.     
More importantly, all those methods represent a semantic class based on a small support set, which is restrictive for capturing rich and fine-grained feature variations required for segmentation tasks. 
  
In this work, we propose a novel prototype-based few-shot learning framework of semantic segmentation to tackle the aforementioned limitations. Our main idea is to enrich the prototype representations of semantic classes in two directions. First, we decompose the commonly used holistic prototype representation into a small set of part-aware prototypes, which is capable of capturing diverse and fine-grained object features and yields better spatial coverage in semantic object regions. Moreover, inspired by the prior work in image classification~\cite{ren2018meta,ayyad2019semi}, we incorporate a set of unlabeled images into our support set so that our part-aware prototypes can be learned from both labeled and unlabeled data source. This enables us to go beyond the restricted small support set and to better model the intra-class variation in object features. We refer to this new problem setting as semi-supervised few-shot semantic segmentation. Based on our new prototypes, we also design a simple and yet flexible matching strategy, which can be applied to either one-way or multi-way setting.   

Specifically, we develop a deep neural network for the task of few-shot semantic segmentation, which consists of three main modules: an embedding network, a prototypes generation network and a part-aware mask generation network. 
Given a few-shot segmentation task, our embedding network module first computes a 2D conv feature map for each image. Taking as input all the feature maps, the prototype generation module extracts a set of part-aware representations of semantic classes from both labeled and unlabeled support images. To achieve this, we first cluster object features into a set of prototype candidates and then use a graph attention network to refine those prototypes using all the support data.
Finally, the part-aware mask generation network fuses the score maps generated by matching
the part-aware prototypes to the query image and predicts the semantic segments.
We train our deep network using the meta-learning strategy with an augmented loss~\cite{yan2019dual} that exploits the original semantic classes for efficient network learning.    
  
We conduct extensive experiments evaluation on the PASCAL-$5^i$\cite{boots2017one,zhang2018sg} and COCO-$20^i$ dataset\cite{wang2019panet,nguyen2019feature} to validate our few-shot semantic segmentation strategy. The results show that our part-aware prototype learning outperforms the state of the art with a large margin. We also include the detailed ablation study in order to provide a better understanding of our method.

The main contribution of this work can be summarized as the following:
\begin{itemize}
	\item We develop a flexible prototype-based method for few-shot semantic segmentation, achieving superior performances in one-way and multi-way setting.
	\item We propose a part-aware prototype representation for semantic classes, capable of encoding fine-grained object features for better segmentation.
	\item To better capture the intra-class variation, we leverage unlabeled data for semi-supervised prototype learning with a graph attention network. 
\end{itemize}

\section{Related Work}

\subsection{Few-shot Classification}
Few-shot learning aims to learn a new concept representation from only a few annotated examples. Most of existing works can be categorized into metric-learning based~\cite{yan2019dual,snell2017prototypical,vinyals2016matching}, optimization-learning based~\cite{ravi2016optimization,finn2017model}, and graph neural network~\cite{garcia2017few,liu2018learning} based methods. Our work is inspired by the metric-learning based methods. In particular, Oriol et al.~\cite{vinyals2016matching} propose to encode an input into an embedded feature and to perform a weighted nearest neighbor matching for classification. The prototypical network~\cite{snell2017prototypical} aims to learn a metric space in which an input is classified according to its distance to class prototypes. Our work is in line with the prototypical network, but we adopt this idea in more challenging segmentation tasks, enjoying a simple design and yet high performance.

There have been several recent attempts aiming to improve the few-shot learning by incorporating a set of unlabeled data, referred to as semi-supervised few-shot learning~\cite{ren2018meta,li2019learning,ayyad2019semi}. Ren et al.~\cite{ren2018meta} first try to leverage unlabeled data to refine the prototypes by Soft $K$-means. Ayyad et al.\cite{ayyad2019semi} introduced a consistency loss both in local and global for utilizing unlabeled data effectively. These methods are initially proposed for solving semi-supervised problems in few-shot classification regime and hence it is non-trivial to extend them to few-shot segmentation directly. We are the first to leverage unlabeled data in the challenging few-shot segmentation task for capturing the large intra-class variations. 

\subsection{Few-shot Semantic Segmentation}
Few-shot semantic segmentation aims to segment semantic objects in an image with only a few annotated examples, and attracted much attention recently. The existing works can be largely grouped into two types: parametric matching-based methods~\cite{zhang2018sg,zhang2019canet,zhang2019pyramid,nguyen2019feature,boots2017one} and prototype-based methods~\cite{siam2019adaptive,wang2019panet}. A recent exception, MetaSegNet~\cite{tian2019differentiable}, adopts the optimization-based few-shot learning strategy and formulates few-shot segmentation as a pixel classification problem.

In the parametric-matching based methods, Shaban et al.~\cite{boots2017one} first develop a weight imprinting mechanism to generate the classification weight for few-shot segmentation.
Zhang et al.~\cite{zhang2019canet} propose to concatenate the holistic objects representation with query features in each spatial position and introduce a dense comparison module to estimate their prediction.The subsequent method, proposed by Zhang et al.~\cite{zhang2019pyramid}, attends to foreground features for each query feature with a graph attention mechanism.
These methods however mainly focus on the restrictive one-way few-shot setting and it is computationally expensive to generalize them to the multi-way setting.

Prototype-based methods conduct pixel-wise matching on query images with holistic prototypes of semantic classes. Wang et al.~\cite{wang2019panet} propose to learn class-specific prototype representation by introducing the prototypes alignment regularization between support and query images. Siam et al.~\cite{siam2019adaptive} adopt a novel multi-resolution prototype imprinting scheme for few-shot segmentation. 
All these prototype-based methods are limited by their holistic representations. To tackle this issue, we propose to decompose object representation into a small set of part-level features for modeling diverse object features at a fine-grained level.

\subsection{Graph Neural Networks}
Our work is related to learning deep networks on graph-structured data. The Graph Neural Networks are first proposed in~\cite{gori2005new,scarselli2008graph} which learn a feature representation via a recurrent message passing process. Graph convolutional networks are a natural generalization of convolutional neural networks to non-Euclidean graphs. Kipf et al.~\cite{kipf2016semi} introduce learning polynomials of the graph laplacian instead of computing eigenvectors to alleviate the computational bottleneck, 
and validated its effectiveness on semi-supervised learning. 
Velic et al.~\cite{velivckovic2017graph} incorporate the attention mechanism into the graph neural network to augment node representation with their contextual information.
Garcia et al.~\cite{garcia2017few} firstly introduce the graph neural network into the few-shot image classification. By contrast, our work employ graph neural network to learn a set of prototypes for the task of semantic segmentation.

\section{Problem Setting}\label{setting}
We consider the problem of few-shot semantic segmentation, which aims to learn to segment semantic objects from only a few annotated training images per class.  
To this end, we adopt a meta-learning strategy~\cite{vinyals2016matching,boots2017one} that builds a meta learner  $\mathcal{M}$ to solve a family of few-shot semantic segmentation tasks $\mathcal{T}=\{T\}$ sampled from an underlying task distribution $P_T$.      

Formally, each few-shot segmentation task $T$ (also called an episode) is composed of a set of support data $\mathcal{S}$ with ground-truth masks and a set of query images $\mathcal{Q}$. 
In our semi-supervised few-shot semantic segmentation setting, the support data $\mathcal{S}=\{\mathcal{S}^l,\mathcal{S}^u\}$ where $\mathcal{S}^l$ and  $\mathcal{S}^u$ are the annotated image-label pairs and unlabeled images, respectively. More specifically, for the $C$-way $K$-shot setting, the annotated support data consists of $K$ image-label pairs from each class, denoted as $\mathcal{S}^l=\{(\mathbf{I}_{c,k}^l, \mathbf{Y}_{c,k}^l)\}_{k=1,...,K}^{c\in \mathcal{C}_T}$, where $\{\mathbf{Y}_{c,k}^l\}$ are pixel-wise annotations, $\mathcal{C}_T$ is the subset of class sets for the task $T$ and $|\mathcal{C}_T|=C$. The unlabeled support images $\mathcal{S}^u=\{\mathbf{I}_i^u\}_{i=1}^{N_u}$ are randomly sampled from the semantic class set $\mathcal{C}$ with their class labels removed during training and inference.
Similarly, the query set $\mathcal{Q}=\{(\mathbf{I}_{j}^q, \mathbf{Y}_{j}^q)\}_{j=1}^{N_q}$, contains $N_q$ images from the class set $\mathcal{C}_T$ whose ground-truth annotations $\{\mathbf{Y}_{j}^q\}$ are provided during training but \textit{unknown} in test.

The meta learner $\mathcal{M}$ aims to learn a functional mapping from the support set $\mathcal{S}$ and a query image $\mathbf{I}^q$ to its segmentation $\mathbf{Y}^q$ for all the tasks. To achieve this, we construct a training set of segmentation tasks $\mathcal{D}^{tr}=\{(\mathcal{S}_n, \mathcal{Q}_n)\}_{n=1}^{|\mathcal{D}^{tr}|}$ with a class set $\mathcal{C}^{tr}$, and train the meta learner episodically on the tasks in $\mathcal{D}^{tr}$. After the meta-training, the model $\mathcal{M}$ encodes the knowledge on how to perform segmentation on different semantic classes across tasks. We finally evaluate the learned model in a test set of tasks $\mathcal{D}^{te}=\{(\mathcal{S}_m,  \mathcal{Q}_m)\}_{m=1}^{|\mathcal{D}^{te}|}$ whose class set $\mathcal{C}^{te}$ is non-overlapped with $\mathcal{C}^{tr}$. 

\begin{figure}[t] 
	\centering
	\includegraphics[width=0.85\linewidth]{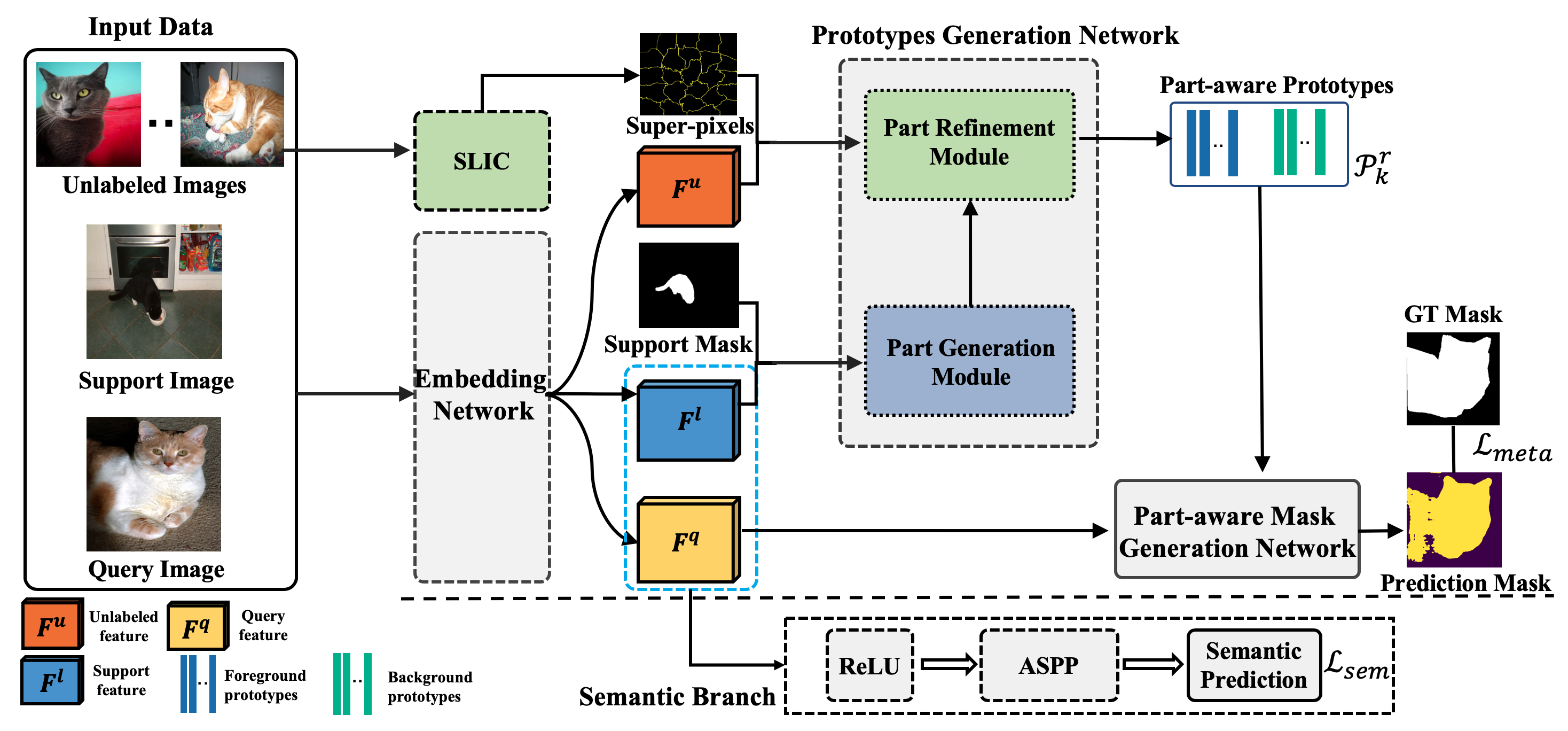}
	\caption{ {\small \textbf{Model Overview:} For each task $T$, \textbf{Embedding Network} first aims to prepare convolutional feature maps of support, unlabeled and query images. \textbf{Prototypes Generation Network} then generates a set of part-aware prototypes by taking support and unlabeled image features as input. It consists of two submodules: \textbf{Part Generation Module} and \textbf{Part Refinement Module} (see below for details). Finally, the \textbf{Part-aware Mask Generation Network} performs segmentation on query features based on a set of part-aware prototypes. In addition, \textbf{Semantic Branch} generates mask predictions over the global semantic class space $\mathcal{C}^{tr}$.}}
	\label{fig-overview}
\end{figure}

\section{Our Approach}
In this work, we adopt a prototype-based few-shot learning framework to build a meta learner $\mathcal{M}$ for semantic segmentation. 
The main idea of our method is to capture the intra-class variation and fine-grained features of semantic classes by a new prototype representation. 
Specifically, we propose to decompose the commonly-used holistic representations of support objects into a set of part-aware prototypes for each class, and additionally utilize unlabeled data to enrich their representations.

To this end, we develop a deep graph network, as our meta learner, to encode such a new representation and to segment the query images. Our network consists of three main networks:  
an \textit{embedding network} that computes convolutional feature maps for the images within a task (in Sec.~\ref{sec:embedding}); 
a \textit{prototypes generation network} that extracts a set of part-aware prototypes from the labeled and unlabeled support images (in Sec.~\ref{sec:meta}); 
and \textit{a part-aware mask generation network} that generates the final semantic segmentation of the query images (in Sec.~\ref{sec:part-aware}).
 
To train our meta model, we adopt a hybrid loss and introduce an auxiliary \textit{semantic branch} that exploits the original semantic classes for efficient learning (in Sec.~\ref{sec:traning}). We refer to our deep model as the \textbf{Part-aware Prototype Network (PPNet)}. An overview of our framework is illustrated in Fig.\ref{fig-overview} and we will introduce the model details in the remaining of this section.

\begin{figure}[t]
    \centering
    \includegraphics[width=0.85\linewidth]{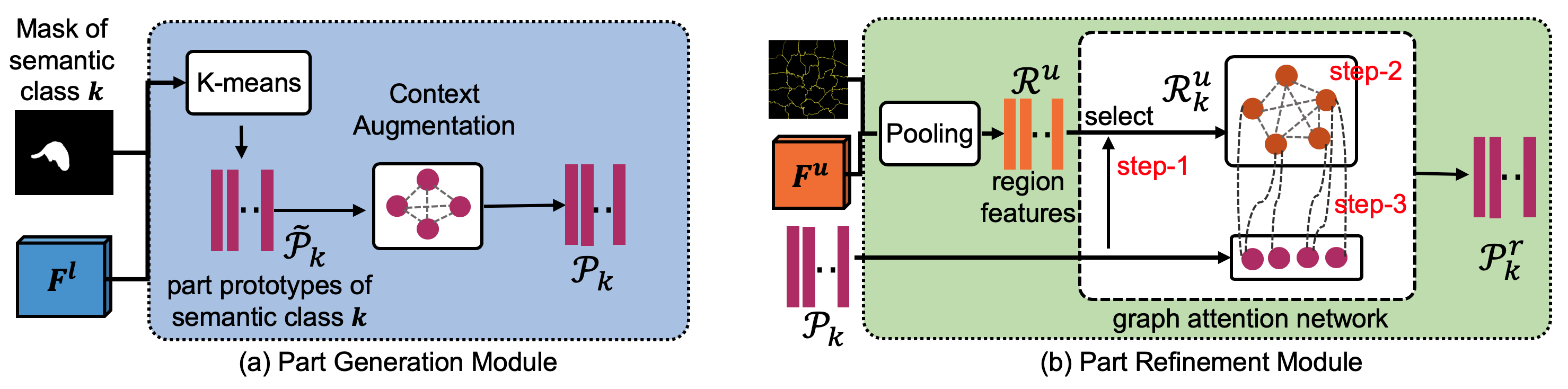}
    \caption{{\small \textbf{Part Generation Module}} aims to generate the initial part-aware prototypes on support images and further incorporate with their global context of the same semantic class. \textbf{Part Refinement Module} further improves part-aware prototypes representation with unlabeled images features by a graph attention network.}
\end{figure}

\subsection{Embedding Network}\label{sec:embedding}
Given a task (or episode), the first module of our PPNet is an embedding network that extracts the convolutional feature maps of all images in the task. Following prior work~\cite{zhang2019canet,zhang2019pyramid}, we adopt ResNet~\cite{he2016deep} as our embedding network, and introduce the dilated convolution to enlarge the receptive field and preserve more spatial details.
Formally, we denote the embedding network as $f_{\text{em}}$, and compute the feature maps of images in a task $T$ as $\mathbf{F} = f_{\text{em}}(\mathbf{I})$, $\forall \mathbf{I}\in \mathcal{S}\cup \mathcal{Q}$. Here $\mathbf{F}\in \mathbb{R}^{H_f\times W_f \times n_{ch} }$, $n_{ch}$ is the number of feature channels, and $(H_f, W_f)$ is the height and width of the feature map. We also resize the annotation mask $\mathbf{Y}$ into the same spatial size as the feature map, denoted as $\mathbf{M}\in \mathbb{R}^{H_f\times W_f}$.

In the $C$-way $K$-shot setting, we reshape and group all the image features in the labeled support set $\mathcal{S}^l$ into $C+1$ subsets: $\mathcal{F}^l=\{\mathcal{F}^l_k, k=0,1,\cdots, C\}$, where $0$ indicates background class and $\mathcal{F}^l_k$ contains all the features $\mathbf{f}\in\mathbb{R}^{n_{ch}}$ annotated with semantic class $k$. 
Similarly, we denote all the features in the unlabeled support set $\mathcal{S}^u$ as $\mathcal{F}^u$.

\subsection{Prototypes Generation Network}\label{sec:meta}
Our second module, the prototypes generation network, aims to generate a set of discriminative part-aware prototypes for each class. For notation clarity, here we focus on a single semantic class $k$. The module takes the image feature set $\mathcal{F}^l_k$ and $\mathcal{F}^u$ as input, and outputs the set of prototypes $\mathcal{P}_k$. 

To this end, we introduce a graph neural network defined on the features, which computes the prototypes in two steps according to the different nature of the labeled and unlabeled support sets. Specifically, the network first extracts part-aware prototypes directly from the labeled support data $\mathcal{F}^l_k$ and then refines their representation by making use of the unlabeled data $\mathcal{F}^u$. As a result, the prototypes generation network consists of two submodules: a \textit{Part Generation Module} and a \textit{Part Refinement Module}, which are described in detail as following.

\subsubsection{Part Generation with Labeled Data} \label{pap-gen}
We first introduce the part generation module, which builds a set of part-aware prototypes from the labeled support set in order to capture fine-grained part-level variation in object regions. 

Specifically, we denote the number of prototypes per class as $N_p$ and the prototype set $\mathcal{P}_k=\{\mathbf{p}_i\}_{i=1}^{N_p}$, $\mathbf{p}_i\in \mathbb{R}^{n_{ch}}$. To define our prototypes, we first compute a data partition $\mathcal{G}=\{G_1,G_2,\cdots,G_{N_p}\}$ on the feature set $\mathcal{F}^l_k$ using the K-means clustering and then generate an initial set of prototypes $\tilde{\mathcal{P}}_k=\{\mathbf{\tilde{p}}_i\}_{i=1}^{N_p}$ with an average pooling layer as follows,   
\begin{align}\label{kmeans-1}
	 \mathbf{\tilde{p}}_i &= \frac{1}{|G_i|}\sum_{j\in G_i} \mathbf{f}_j, 
	 \quad \mathbf{f}_j\in\mathcal{F}^l_k 
\end{align}
We further incorporate a global context of the semantic class into the part-aware prototypes by augmenting each initial prototype with a context vector, which is estimated from other prototypes in the same class based on the attention mechanism~\cite{vaswani2017attention}:  
\begin{align}
    \label{gst}
     \mathbf{{p}}_i=  \mathbf{\tilde{p}}_i + \lambda_p \sum_{\substack{j=1\wedge j\neq i}}^{N_p} \mu_{ij}\mathbf{\tilde{p}}_j,\quad
     \mu_{ij}=\frac{d(\mathbf{\tilde{p}}_i, \mathbf{\tilde{p}}_j)}{\sum_{j\neq i} d(\mathbf{\tilde{p}}_i, \mathbf{\tilde{p}}_j) }
\end{align}
where $\lambda_p$ is a scaling parameter and $\mu_{ij}$ is the attention weight calculated with a similarity metric $d$, e.g., cosine similarity. 

\subsubsection{Part Refinement with Unlabeled Data:}
The second submodule, the part refinement module, aims to capture the intra-class variation of each semantic class by enriching the prototypes on additional unlabeled support images. However, exploiting the unlabeled data is challenging due to the fact that the set of unannotated image features $\mathcal{F}^u$ is much more noisy and in general has a much larger volume than the labeled set $\mathcal{F}_k^l$.

We tackle the above two problems by a grouping and pruning process, which yields a smaller and more relevant set of features $\mathcal{R}^u_k$ for class $k$. Based on $\mathcal{R}^u_k$, we then design a graph attention network to smooth the unlabeled features and to refine the part-aware prototypes by aggregating those features.   
Concretely, our refinement process includes the following three steps: 

Step-1: \textit{Relevant feature generation.} 
We first compute a region-level feature representation of unlabeled images by utilizing the idea of superpixel generation. Concretely, we apply SLIC~\cite{alkslic} to all the unlabeled images and generate a set of groupings on $\mathcal{F}^u$. Denoting the groupings as $\mathcal{R}=\{R_1,R_2,\cdots, R_{N_r}\}$, we use the average pooling to produce a pool of region-level features $\mathcal{R}^u=\{\mathbf{r}_i\}_{i=1}^{N_r}$. We then select a set of relevant features for class $k$ as follows:
\begin{align}
 \mathcal{R}^u_k = \{\mathbf{r}_j: \mathbf{r}_j \in \mathcal{R}^u \wedge \exists \mathbf{p}_i \in \mathcal{P}_k,  d({\mathbf{p}}_i, \mathbf{r}_j) >\sigma\}
\end{align}
where $d(\cdot,\cdot)$ is a {cosine similarity} function between prototype and feature, and $\sigma$ is a threshold that determines the relevance of the features.  

Step-2: \textit{Unlabeled feature augmentation.}
With the selected unlabeled features, the second step aims to enhance those region-level representations by incorporating contextual information in the unlabeled feature set. This allows us to encode both local and global cues of a semantic class.  

Specifically, we build a fully-connected graph on the feature set $\mathcal{R}^u_k$ and use the following message passing function to compute the update $\tilde{\mathcal{R}}^u_k=\{\tilde{\mathbf{r}}_i\}_{i=1}^{|\tilde{\mathcal{R}}^u_k|}$: 
\begin{align}
\label{inner}
{\tilde{\mathbf{r}}}_i=\mathbf{r}_i + h\left(\frac{1}{Z_i^u}\sum_{j=1\wedge j\neq i}^{|\mathcal{R}^u_k|} d(\mathbf{r}_i,\mathbf{r}_j)\mathbf{W}\mathbf{r}_j\right)
\end{align}
where ${\tilde{\mathbf{r}}}_i$ represents the updated representation at node $i$, $h$ is an element-wise activate function (e.g., ReLU). $d$ is a similarity function encoding the relations between two feature vectors $\mathbf{r}_i$ and $\mathbf{r}_j$, and $Z_i^u$ is a normalization factor for node $i$. $\mathbf{W}\in \mathbb{R}^{n_{ch}\times n_{ch}}$ is the weight matrix defining a linear mapping to encode the message from node $j$. 

Step-3: \textit{Part-aware prototype refinement.}
Given the augmented unlabeled features, we refine the original part-aware prototypes with an attention strategy similar to the labeled one.
We use the part-aware prototypes $\mathcal{P}_k$ as attention query to choose  similar unlabeled features in $\tilde{\mathcal{R}}^u_k$ and aggregate them into $\mathcal{P}_k$: 
\begin{align}
\label{inter}
     \mathbf{p}^{r}_i= {\mathbf{p}}_i + \lambda_r \sum_{j=1}^{|\tilde{\mathcal{R}}^u_k|} \phi_{ij}{\tilde{\mathbf{r}}}_j,\quad
     \phi_{ij}=\frac{d({\mathbf{p}}_i, {\tilde{\mathbf{r}}}_j)}{\sum_{j} d({\mathbf{p}}_i, {\tilde{\mathbf{r}}}_j)}
\end{align}
where $\lambda_r$ is a scaling parameter and $\phi_{ij}$ is the attention weight. The final refined prototype set for class $k$ is denoted as $\mathcal{P}_k^{r}=\{ \mathbf{p}^{r}_1, \mathbf{p}^{r}_2,\cdots, \mathbf{p}^{r}_{N_p}\}$.

\subsection{Part-aware Mask Generation Network}\label{sec:part-aware}
Given the part-aware prototypes $\{\mathcal{P}^{r}_k\}_{k=0}^C$ of every semantic class and background in each task, we introduce a simple and yet flexible matching strategy to generate the semantic mask prediction on a query image $\mathbf{I}^q$. We denote its conv feature map as $\mathbf{F}^q$ and its feature column at location $(m,n)$ as $\mathbf{f}^q_{m,n}$.

Specifically, we first generate a similarity score map for each part-aware prototype performing the \textit{prototype-pixel} matching as follows, 
\begin{align}
\mathbf{S}_{k,j}(m,n) = d(\mathbf{f}^q_{m,n}, \mathbf{p}^{r}_j), \quad \mathbf{p}^{r}_j\in\mathcal{P}^{r}_k,\quad \mathbf{S}_{k,j}\in \mathbb{R}^{H_f\times W_f}
\end{align}
where $d$ is the {cosine similarity} function and $\mathbf{S}_{k,j}(m,n)$ is the score at location $(m,n)$. We then fuse together all the score maps from the class $k$ by max-pooling and generate the output segmentation scores by concatenating score maps from all the classes: 
\begin{align}
\mathbf{S}_k^q =\text{MaxPool}(\{\mathbf{S}_{k,j}\}_{j=1}^{N_p}),\quad
\mathbf{\hat{Y}}^q = \bigoplus\{\mathbf{S}_k^q\}_{k=0}^C
\end{align}
where $\bigoplus$ indicates concatenation. To generate the final segmentation, we upsample the score output $\mathbf{\hat{Y}}^q$ by bilinear interpolation and choose the class label with the highest score at each pixel location.

\subsection{Model Training with Semantic Regularization}\label{sec:traning}
To estimate the parameters of proposed model, we train our PPNet in the meta-learning framework. Specifically, we adopt the standard cross-entropy loss to train our entire network on all the tasks in the training set $\mathcal{D}^{tr}$. Inspired by~\cite{wang2019panet}, we compute the cross-entropy loss on both support and query images. The loss for each task can be written as:

\begin{equation}
    \mathcal{L}_{meta} = \mathcal{L}_{CE}(\hat{\mathbf{Y}}^q, \mathbf{Y}^q) + \mathcal{L}_{CE}(\hat{\mathbf{Y}}^l, \mathbf{Y}^l)
\end{equation}
where $\mathcal{L}_{CE}$ is the cross-entropy function, $\mathbf{Y}^l, \mathbf{\hat{Y}}^{l}$ are the ground-truth and prediction mask for support image. We note that while our full model is not strictly differentiable w.r.t the embedding network thanks to the prototype clustering and candidate region selection, we are able to compute an approximate gradient by fixing the clustering and selection outcomes. This approximation works well empirically largely due to a well pre-trained embedding network.

In order to learn better visual representation for few-shot segmentation, we introduce another \textbf{semantic branch}~\cite{yan2019dual} for computing a semantic loss defined on the global semantic class space $C^{tr}$ (in contrast to $C$ classes in individual tasks). To achieve this, we augment the network with an Atrous Spatial Pyramid Pooling module (ASPP) decoder to predict mask scores $\hat{\mathbf{Y}}^{q}_{sem},\hat{\mathbf{Y}}^l_{sem}$ of support and query image respectively in the global class space $C^{tr}$, and compute the semantic loss as below, 
\begin{equation}
    \mathcal{L}_{sem} = \mathcal{L}_{CE}(\hat{\mathbf{Y}}_{sem}^q, \mathbf{Y}_{sem}^q) + \mathcal{L}_{CE}(\hat{\mathbf{Y}}^l_{sem}, \mathbf{Y}_{sem}^{l})
\end{equation}
Here $\mathbf{Y}_{sem}^q, \mathbf{Y}_{sem}^{l}$ are ground-truth masks defined over shared class space $C^{tr}$. The overall training loss for each task is:
\begin{equation}
    \mathcal{L}_{full} = \mathcal{L}_{meta} +\beta \mathcal{L}_{sem}
\end{equation}
where $\beta$ is hyper-parameter to balance the weight of task loss and semantic loss.

\section{Experiments}
We evaluate our method on the task of few-shot semantic segmentation by conducting a set of experiments on two datasets, including PASCAL-$5^i$~\cite{boots2017one} and COCO-$20^i$~\cite{wang2019panet,nguyen2019feature}. In each dataset, we compare our approach with the state-of-the-art methods in terms of prediction accuracy.

Below we first introduce the implementation details and experimental configuration in Sec.~\ref{sec:config}. Then we present our experimental analysis on PASCAL-$5^i$ dataset in Sec.~\ref{sec:pascal}, followed by our results on the COCO-$20^i$ dataset in Sec.~\ref{sec:coco}. We report comparisons of quantitative results and analysis on each dataset. Finally, we conduct a series of ablation studies to evaluate the importance of each component of the model in Sec.~\ref{sec:ablation}.

\subsection{Experimental Configuration}\label{sec:config}
\paragraph{{\rm \textbf{Network Details:}}}
We adopt ResNet~\cite{he2016deep}, initialized with weights pre-trained on ILSVRC~\cite{russakovsky2015imagenet}, as feature extractor to compute the convolutional feature maps. In last two res-blocks, the strides of max-pooling are set as 1 and dilated convolutions are taken with dilation rate 2, 4 respectively. The last ReLU operation is removed for generating the prototypes. The input images are resized into a fixed resolution [417,417] and horizontal random flipping is used for data augmentation. For the part-aware prototypes network, the typical hyper-parameter of the parts is $N_p=5$. In the part refinement module, we first generate $N_r$=100 candidate regions on unlabeled data, and select the relevant regions for each semantic class by setting similarity threshold $\sigma$ as 0. In addition,  $\lambda_p$ in Eq.~\ref{gst} and $\lambda_r$ in Eq.~\ref{inter} are set to $0.8$ and $0.2$ respectively, which control the proportion of parts and unlabeled information passed.

\paragraph{{\rm \textbf{Training Setting:}}} 
For the meta-training phase, the model is trained with the SGD optimizer, initial learning rate 5e-4, weight decay 1e-4 and momentum 0.9. We train 24k iterations in total, and decay the learning rate 10 times in 10k, 20k iteration respectively. The weight $\beta$ of semantic loss $\mathcal{L}_{sem}$ is set as 0.5. At the testing phase, we average the mean-IoU of 5-runs~\cite{wang2019panet} with different random seeds in each fold with each run containing 1000 tasks. 

\paragraph{{\rm \textbf{Baseline and Evaluation Metrics:}}} We adopt ResNet-50~\cite{he2016deep} as feature extractor in PANet~\cite{wang2019panet} to be our baseline model, denoted as PANet*. 
Following previous works~\cite{boots2017one,zhang2018sg,rakelly2018conditional,wang2019panet,zhang2019pyramid}, mean-IoU and binary-IoU are adopted for model evaluation.  Mean-IoU measures the averaged Intersection-over-Union (IoU) of all the classes. Binary-IoU\footnote{We report Binary-IoU in supplementary material for a clear comparison with the previous works.}is calculated by treating all object classes as the foreground and averaging the IoU of foreground and background. In our experiments, we mainly focus on mean-IoU metrics for evaluation since it reflects the model generalization ability more precisely.


\begin{table*}[t]
	\caption{{\small Mean-IoU of 1-way on PASCAL-$5^i$. $*$ denotes the results implemented by ourselves. MS denotes the model evaluated with multi-scale inputs.\cite{zhang2019pyramid,zhang2019canet}. Red numbers denote the averaged mean-IoU over 4 folds.}}
	\centering
	\label{w1-pascal}
	\resizebox{0.8\textwidth}{!}{
		\begin{tabular}{c|c|c|ccccc|ccccc|c}
			\hline
			\hline
			\multirow{2}{*}{Methods}&\multirow{2}{*}{MS}&\multirow{2}{*}{Backbone}&\multicolumn{5}{c|}{\textbf{1-shot}}&\multicolumn{5}{c|}{\textbf{5-shot}} &\multirow{2}{*}{\#params}\\
			&&&fold-1&fold-2 &fold-3&fold-4 &{\color[rgb]{0.7725,0.3529,0.06667}mean}&fold-1&fold-2 &fold-3&fold-4 &{\color[rgb]{0.7725,0.3529,0.06667}mean} & \\
			\hline
			OSLSM~\cite{boots2017one}              &x&VGG16       &33.60&55.30&40.90&33.50&{\color[rgb]{0.7725,0.3529,0.06667}40.80}  &35.90&58.10&42.70&39.10&{\color[rgb]{0.7725,0.3529,0.06667} 43.90} &272.6M\\
			co-FCN~\cite{rakelly2018conditional}   &x&VGG16       &31.67&50.60&44.90&32.40&{\color[rgb]{0.7725,0.3529,0.06667}41.10}  &37.50&50.00&44.10&33.90&{\color[rgb]{0.7725,0.3529,0.06667} 41.40} &34.20M\\
			SG-one~\cite{zhang2018sg}              &x&VGG16       &40.20&58.40&48.40&38.40&{\color[rgb]{0.7725,0.3529,0.06667}46.30}  &41.90&58.60&48.60&39.40&{\color[rgb]{0.7725,0.3529,0.06667} 47.10} &19.00M\\
			AMP~\cite{siam2019adaptive}            &x&VGG16       &36.80&51.60&46.90&36.00&{\color[rgb]{0.7725,0.3529,0.06667}42.80}  &44.60&58.00&53.30&42.10&{\color[rgb]{0.7725,0.3529,0.06667}49.50} &15.8M\\
			PANet~\cite{wang2019panet}             &x&VGG16       &42.30&58.00&51.10&41.20&{\color[rgb]{0.7725,0.3529,0.06667}48.10}  &51.80&64.60&59.80&46.50&{\color[rgb]{0.7725,0.3529,0.06667}55.70} &14.7M\\
			PANet*~\cite{wang2019panet}            &x&RN50        &44.03&57.52&50.84&44.03&{\color[rgb]{0.7725,0.3529,0.06667}49.10}  &55.31&67.22&61.28&53.21&{\color[rgb]{0.7725,0.3529,0.06667}59.26} &23.5M\\
			PGNet*~\cite{zhang2019pyramid}         &x&RN50        &53.10&63.60&47.60&47.70&{\color[rgb]{0.7725,0.3529,0.06667}53.00}  &56.30&66.10&48.00&53.20&{\color[rgb]{0.7725,0.3529,0.06667}55.90} &32.5M\\
			FWB\cite{nguyen2019feature}            &x&RN101       &51.30&64.49&56.71&\textbf{52.24}&{\color[rgb]{0.7725,0.3529,0.06667}\textbf{56.19}}  &54.84&67.38&62.16&55.30&{\color[rgb]{0.7725,0.3529,0.06667}59.92} &43.0M\\ 
			\hline
			CANet~\cite{zhang2019canet}    &\checkmark&RN50       &52.50&65.90&51.30&51.90&{\color[rgb]{0.7725,0.3529,0.06667}55.40}  &55.50&67.80&51.90&53.20&{\color[rgb]{0.7725,0.3529,0.06667}57.10} &36.35M\\ 
			PGNet~\cite{zhang2019pyramid}     &\checkmark&RN50    &\textbf{56.00}&\textbf{66.90}&50.60&50.40&{\color[rgb]{0.7725,0.3529,0.06667}56.00}  &57.70&68.70&52.90&54.60&{\color[rgb]{0.7725,0.3529,0.06667}58.50} &32.5M\\ 
			\hline
			Ours(w/o \textbf{$\mathcal{S}^u$})      &x&RN50        &47.83&58.75&53.80&45.63&{\color[rgb]{0.7725,0.3529,0.06667}51.50}  &58.39&67.83&64.88&56.73&{\color[rgb]{0.7725,0.3529,0.06667}61.96} &23.5M\\
			our                                    &x&RN50        &48.58&60.58&55.71&46.47&{\color[rgb]{0.7725,0.3529,0.06667}52.84}  &\textbf{58.85}&\textbf{68.28}&\textbf{66.77}&\textbf{57.98}&{\color[rgb]{0.7725,0.3529,0.06667}\textbf{62.97}} &31.5M\\
			Ours                                    &x&RN101       &52.71&62.82&\textbf{57.38}&47.74&{\color[rgb]{0.7725,0.3529,0.06667}55.16}  &\textbf{60.25}&\textbf{70.00}&\textbf{69.41}&\textbf{60.72}&{\color[rgb]{0.7725,0.3529,0.06667}\textbf{65.10}}&50.5M \\
			\hline
			\hline
	\end{tabular}}
\end{table*}
	
\begin{table*}[ht]
	\center
	\caption{{\small Mean-IoU of 2-way on PSACAL-$5^i$. $*$ denotes our implementation. Red numbers denote the averaged mean-IoU over 4 folds.}}\label{w2-para}
	\resizebox{0.8\textwidth}{!}{
		\begin{tabular}{c|c|ccccc|ccccc}
			\hline
			\hline
			\multirow{2}{*}{Methods}&\multirow{2}{*}{Backbone}&\multicolumn{5}{c|}{\textbf{1-shot}}&\multicolumn{5}{c}{\textbf{5-shot}}\\
			&&fold-1&fold-2 &fold-3&fold-4 &{\color[rgb]{0.7725,0.3529,0.06667}mean}&fold-1&fold-2 &fold-3&fold-4 &{\color[rgb]{0.7725,0.3529,0.06667}mean}\\
			\hline
			\hline
			MetaSegNet~\cite{tian2019differentiable}      &RN9&-&-&-&-&-             &41.9&41.75&46.31&43.63&{\color[rgb]{0.7725,0.3529,0.06667}43.30}\\
			
			PANet\cite{wang2019panet}                     &VGG16&-&-&-&-&{\color[rgb]{0.7725,0.3529,0.06667}45.1}        &-&-&-&-&{\color[rgb]{0.7725,0.3529,0.06667}53.10} \\
			PANet*\cite{wang2019panet}                    &RN50&42.82&56.28&48.72&45.53&{\color[rgb]{0.7725,0.3529,0.06667}48.33}  &54.65&64.80&57.61&54.94&{\color[rgb]{0.7725,0.3529,0.06667}58.00} \\
			\hline
			Ours(\textbf{w/o $\mathcal{S}^u$)}           &RN50&45.63&58.00&51.65&45.69&{\color[rgb]{0.7725,0.3529,0.06667}50.24}  &55.34&66.38&63.79&56.85&{\color[rgb]{0.7725,0.3529,0.06667}60.59} \\
			Ours                                         &RN50&\textbf{47.36}&\textbf{58.34}&\textbf{52.71}&\textbf{48.18}&{\color[rgb]{0.7725,0.3529,0.06667}\textbf{51.65}}  &\textbf{55.54}&\textbf{67.26}&\textbf{64.36}&\textbf{58.02}&{\color[rgb]{0.7725,0.3529,0.06667}\textbf{61.30}}\\
			\hline
			\hline
	\end{tabular}}
\end{table*}

\subsection{Experiments on PASCAL-$5^i$}\label{sec:pascal}
\paragraph{{\rm\textbf{Dataset:}}}
The PASCAL-$5^i$ is introduced in~\cite{boots2017one,zhang2018sg}, which is created from PASCAL VOC 2012 
dataset with SBD~\cite{hariharan2011semantic} augmentation. Specifically, the 20 classes in  PASCAL VOC is split into 4-folds evenly, each containing 5 categories. Models are trained on three folds and evaluated on the rest using cross-validation.

\begin{figure}[ht]
  \centering
  \includegraphics[width=0.85\linewidth]{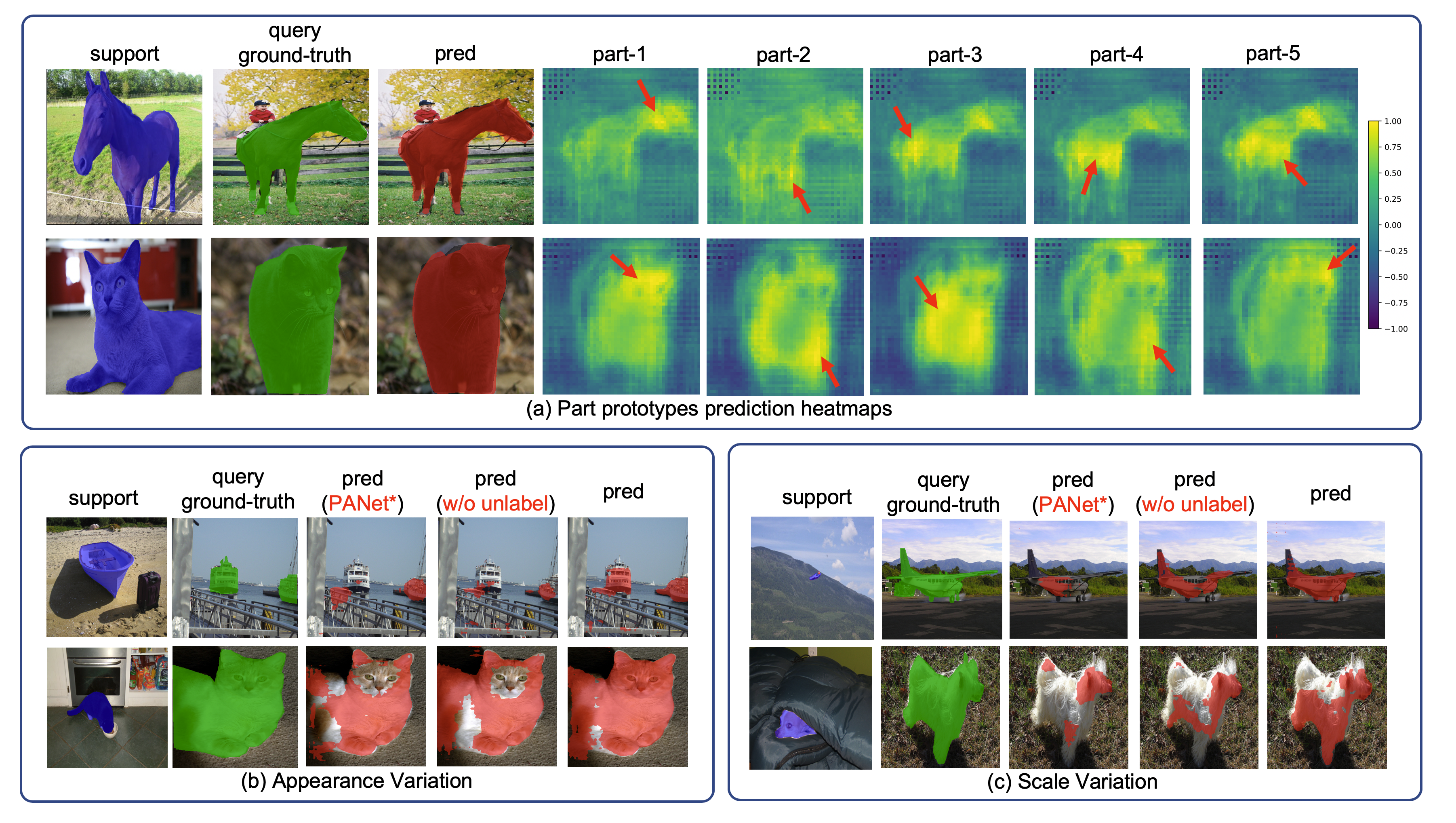}
  \caption{{\small Qualitative Visualization of 1-way 1-shot setting on PASCAL-$5^i$}. (a) demonstrates the part-aware prototypes response heatmaps. The bright region denotes a high similarity between prototypes and query images. (b) and (c) show the capabilities of our model in coping with appearance and scale variation by utilizing unlabeled data. {\color{red} Red masks} denote prediction results of our models. {\color{blue} Blue} and {\color{green}green} masks denote the ground-truth of support and query images. See suppl. for more visualization results.}
  \label{demo} 
\end{figure}
\paragraph{{\rm\textbf{Quantitative Results:}}} 
We compare the performance of our PPNet with the previous state-of-the-art methods. The detail results of 1-way setting are reported in Tab.\ref{w1-pascal}. With the ResNet-50 as the embedding network, our model achieves \textbf{61.96}\% mean-IoU in 5-shot setting, which outperforms the state-of-the-art method with a sizable margin of \textbf{2.71}\%. The performance can be further improved to \textbf{52.84}\%(1-shot) and \textbf{62.97}\%(5-shot) by refining the part prototypes with the unlabeled data. Compared with the PANet~\cite{wang2019panet}, our method achieves the considerable improvement in both 1-shot and 5-shot setting, which demonstrates the part-aware prototypes can provide a superior representation than the holistic prototype. 
Moreover, our method achieves the state-of-the-art performance at \textbf{65.10}\% in 5-shot setting relied on the ResNet-101 backbone\footnote{We note that our 1-shot performance is affected by the limited representation power of the prototypes learned from a single support image while prior methods~\cite{zhang2019pyramid,zhang2019canet} employ a complex Convnet decoder to exploit additional spatial smoothness prior.}.

To investigate the effectiveness of our method on multi-way setting, a 2-way experiment is conducted and the results are reported in Tab.\ref{w2-para}. Our method can outperform the previous works both in with/without unlabeled data with a large margin. Our model can achieve \textbf{51.65}\% and \textbf{61.30}\% for 1-shot and 5-shot setting, which has \textbf{3.32}\% and \textbf{3.30}\% performance gain compared with PANet*, respectively. The quantitative results indicate our PPNet can achieve the state-of-the-art in a more challenging setting.

\paragraph{{\rm\textbf{Visualization Analysis:}}} 
To better understand our part-aware prototype framework, we visualize the responding region of our part prototypes and the prediction results in Fig.\ref{demo}.
The response heatmaps are presented in the column 4-8 of the (a). For example, given a support image(horse or cat), the part prototypes are corresponding to different body parts, which is capable of modeling one semantic class at a fine-grained level.

Moreover, our model can cope with the large appearance and scale variation between support and query images, which is illustrated in (b) and (c). Compared with the PANet*, our method can enhance the modeling capability with the part prototypes, and has a significant improvement on the segmentation prediction by utilizing the unlabeled images to better model the intra-class variations.

\subsection{Experiments on COCO-$20^i$}\label{sec:coco}
\paragraph{{\rm \textbf{Dataset:}}} COCO-$20^i$~\cite{wang2019panet,nguyen2019feature} is another more challenging benchmark built from MSCOCO~\cite{lin2014microsoft}. Similar to PASCAl-$5^i$, MS-COCO dataset is divided into 4-folds with 20 categories in each fold. There are two splits of MSCOCO: we refer the data partition in \cite{wang2019panet} as \textit{split-A} while the split in \cite{nguyen2019feature} as \textit{split-B}. We mainly focus on \textit{split-A} and also report the performance on \textit{split-B}.  Models are trained on three folds and evaluated on the rest with a cross-validation strategy. 

\begin{table*}[t]
  \center
  \caption{{\small Mean-IoU results of 1-way on COCO-$20^i$ split-A. Red numbers denote the averaged mean-IoU over 4 folds. * is our implementation}}
  \label{w1-coco}
  \resizebox{0.8\textwidth}{!}{
  \begin{tabular}{c|c|c|ccccc|ccccc}
  \hline
  \hline
  \multirow{2}{*}{Methods}&\multirow{2}{*}{Split}&\multirow{2}{*}{Backbone}&\multicolumn{5}{c|}{\textbf{1-shot}}&\multicolumn{5}{c}{\textbf{5-shot}}
  \\
  &&&fold-1&fold-2 &fold-3&fold-4 &{\color[rgb]{0.7725,0.3529,0.06667}mean}&fold-1&fold-2 &fold-3&fold-4 &{\color[rgb]{0.7725,0.3529,0.06667}mean}\\
  \hline
  \hline
  PANet~\cite{wang2019panet}            &A&VGG16   &28.70&21.20&19.10&14.80&{\color[rgb]{0.7725,0.3529,0.06667}20.90}  &39.43&28.30&28.20&22.70&{\color[rgb]{0.7725,0.3529,0.06667}29.70} \\
  PANet*~\cite{wang2019panet}           &A&RN50    &31.50&22.58&21.50&16.20&{\color[rgb]{0.7725,0.3529,0.06667}22.95}  &45.85&29.15&30.59&29.59&{\color[rgb]{0.7725,0.3529,0.06667}33.80}\\
  Ours(\textbf{w/o $\mathcal{S}^u$})     &A&RN50  &34.53&25.44&24.33&18.57&{\color[rgb]{0.7725,0.3529,0.06667}25.71}  &48.30&30.90&35.65&30.20&{\color[rgb]{0.7725,0.3529,0.06667}36.24}\\
  Ours                                   &A&RN50    &\textbf{36.48}&\textbf{26.53}&\textbf{25.99}&\textbf{19.65}&{\color[rgb]{0.7725,0.3529,0.06667}\textbf{27.16}}  &\textbf{48.88}&\textbf{31.36}&\textbf{36.02}&\textbf{30.64}&{\color[rgb]{0.7725,0.3529,0.06667}\textbf{36.73}}\\
  \hline
  FWB~\cite{nguyen2019feature}          &B&RN101   &16.98&17.78&20.96&\textbf{28.85}&{\color[rgb]{0.7725,0.3529,0.06667}21.19}  &19.13&21.46&23.39&30.08&{\color[rgb]{0.7725,0.3529,0.06667}23.05} \\
  Ours                                   &B&RN50    &\textbf{28.09}&\textbf{30.84}&\textbf{29.49}&27.70&{\color[rgb]{0.7725,0.3529,0.06667}\textbf{29.03}}  &\textbf{38.97}&\textbf{40.81}&\textbf{37.07}&\textbf{37.28}&{\color[rgb]{0.7725,0.3529,0.06667}\textbf{38.53}}\\
  \hline
  \hline
  \end{tabular}}
\end{table*}

\begin{table*}[t]
  \center
  \caption{{\small Ablation Studies of 1-way 1-shot on COCO-$20^i$ split-A in every fold. Red numbers denote the averaged mean-IoU over 4 folds.}}
  \label{aba}
  \resizebox{0.7\textwidth}{!}{
          \begin{tabular}{c|ccc|ccccc}
              \hline
              \hline
              &&&&\multicolumn{5}{c}{\textbf{1-shot}}
              \\
              Model                                &PAP&SEM&UD &fold-1&fold-2 &fold-3&fold-4 &{\color[rgb]{0.7725,0.3529,0.06667}mean}\\
              \hline
              \hline
              Baseline (PANet*)                    &-&-&-&31.50&22.58&21.50&16.20&{\color[rgb]{0.7725,0.3529,0.06667}22.95} \\
                                                   &\checkmark(w/o context)&-&-&32.05&23.09&21.33&16.94&{\color[rgb]{0.7725,0.3529,0.06667}23.35} \\
                                                   &\checkmark&-&-&34.45&24.37&23.46&17.79&{\color[rgb]{0.7725,0.3529,0.06667}25.02} \\
                                                   &\checkmark&\checkmark&-&34.53&25.44&24.33&18.57&{\color[rgb]{0.7725,0.3529,0.06667}25.71} \\
                                                   &\checkmark&\checkmark&\checkmark(w/o \textit{step-2}) &36.02&24.45&25.82&19.07&{\color[rgb]{0.7725,0.3529,0.06667}26.34} \\
              Ours                                &\checkmark&\checkmark&\checkmark&\textbf{36.48}&\textbf{26.53}&\textbf{25.99}&\textbf{19.65}&{\color[rgb]{0.7725,0.3529,0.06667}\textbf{27.16}} \\
              \hline
              \hline
              \end{tabular}}
\end{table*}

\paragraph{{\rm \textbf{Quantitative Results:}}} 
We report the performance of our method on this more challenging benchmark in Tab.\ref{w1-coco}. Compared with the recent works~\cite{wang2019panet,nguyen2019feature}, our method can achieve  the state-of-the-art performance in different splits that use the same type of embedding networks by a sizable margin. Compared with the baseline PANet*, the same performance improvement trends are shown in both setting. In split-B~\cite{nguyen2019feature}, our model is superior to FWB~\cite{nguyen2019feature} nearly in every fold, except for fold-4 in 1-shot, achieving 29.03\% in 1-shot and 38.53\% in 5-shot.

\subsection{Ablation Study}\label{sec:ablation}
In this subsection, we conduct several experiments to evaluate the effectiveness of our model components on COCO-$20^i$ \textit{split}-A 1-way 1-shot setting.

\subsubsection{{\rm\textbf{Part-aware Prototypes (PAP):}}}As in Tab.~\ref{aba}, by decomposing the holistic object representation~\cite{wang2019panet,zhang2018sg,zhang2019canet} into a small set of part-level representations, the averaged mean-IoU is improved from 22.95\% to \textbf{23.35\%}. We further demonstrate the effectiveness of the global context used for augmenting part prototype(in Eq.\ref{gst}). The performance can achieve continuous improvement to 25.02\%, which suggests that global semantic is important for part-level representations.

\subsubsection{{\rm \textbf{Semantic Branch (SEM):}}} We also conduct experiments to validate the semantic branch~\cite{yan2019dual}. It is evident that the semantic branch is able to improve the convergence and the final performance significantly, which indicates that the full PPNet exploits the semantic information efficiently.

\subsubsection{{\rm \textbf{Unlabel Data (UD):}}} We also investigate the graph attention network for the exploitation of the unlabeled data. As discussed in the method, we propose to utilize the graph attention network to refine the part prototypes. We compare the performance of the full PPNet with the PPNet without the GNN module used in \textit{step-2}. The performance demonstrates the effectiveness of the GNN, and our full PPNet can achieve 27.16\% in terms of averaged mean-IoU over 4 folds.

\begin{figure}[t]
    \centering
    \includegraphics[width=0.85\linewidth]{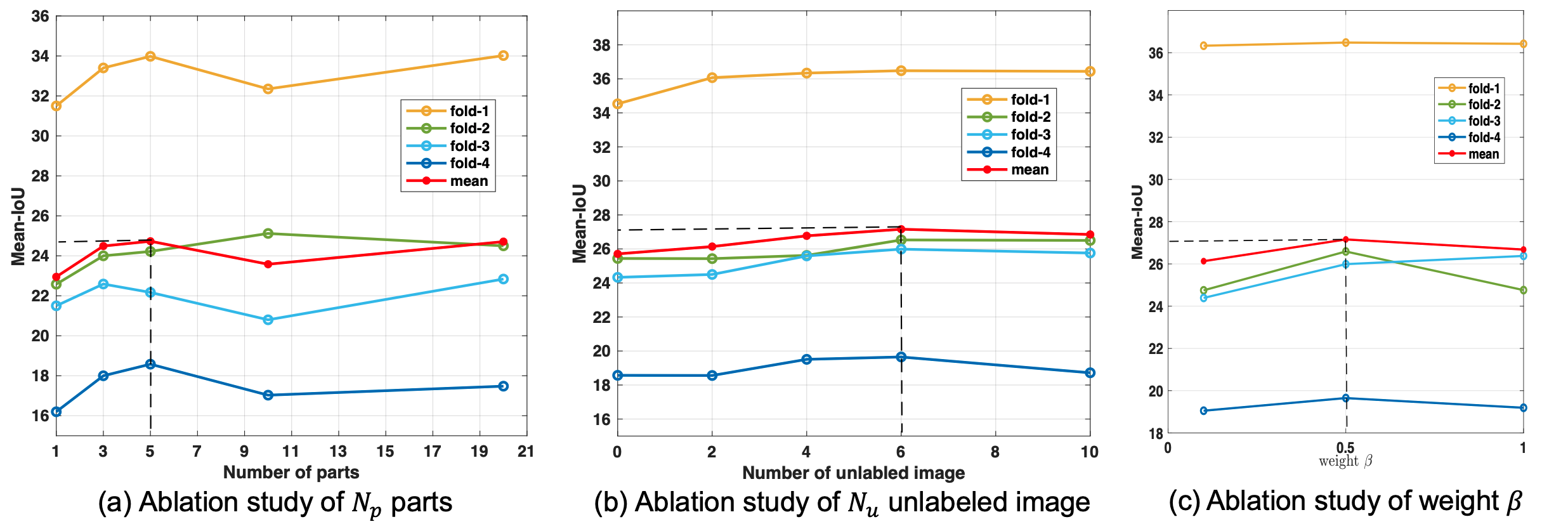}
    \caption{{\small Ablation studies of $N_p$ parts, $N_u$ unlabeled data and weight $\beta$ for semantic loss on COCO-$20^i$ split-A 1-way 1-shot.}}
    \label{aba-ku}
\end{figure}

\paragraph{{\rm \textbf{Hyper-parameters $N_p$, $N_u$ and $\beta$:}}} 
We conduct several ablation studies to explore the influence of the hyper-parameters of our PPNet. We first investigate the part number $N_p$ on `baseline+PAP' model and plot the performance curve in Fig.\ref{aba-ku}(a). In our experiments, the highest performance is achieved when $N_p$ is 5 and 20 (red line) over 4 folds, and we set $N_p$=5 for computation efficiency. In our semi-supervised few-shot segmentation task, we also investigate the influence of the unlabeled image number $N_u$. In Fig.\ref{aba-ku}(b), we can achieve the highest averaged mean-IoU over 4 folds (red line) with our full PPNet when $N_u$=6. In addition, we also investigate the weight $\beta$ for semantic loss $\mathcal{L}_{sem}$ in our final model during training stage. As shown in Fig.\ref{aba-ku}(c), the optimal value is $\beta$=0.5.

\section{Conclusion}
In this work, we presented a flexible prototype-based method for few-shot semantic segmentation. Our approach is able to capture diverse appearances of each semantic class. To achieve this, we proposed a part-aware prototypes representation to encode the fine-grained object features. In addition, we leveraged unlabeled data to capture the intra-class variations of the prototypes, where we introduce the first framework of semi-supervised few-shot semantic segmentation. We developed a novel graph neural network model to generate and enhance the part-aware prototypes based on support images with and without pixel-wise annotations. We evaluated our method on several few-shot segmentation benchmarks, in which our approach outperforms the prior works with a large margin, achieving the state-of-the-art performance.

%
%
\bibliographystyle{splncs04}
\bibliography{egbib}
\end{document}


\pagestyle{headings}
\mainmatter
\def\ECCVSubNumber{754}  

\title{Part-aware Prototype Network for Few-shot Semantic Segmentation\\Supplementary Material} 

\titlerunning{Part-aware Prototype Network for Few-shot Semantic Segmentation} 
\author{Yongfei Liu\textsuperscript{\rm 1}\thanks{Both authors contributed equally to the work. This work was supported by Shanghai NSF Grant (No. 18ZR1425100)} \and
Xiangyi Zhang\textsuperscript{\rm 1}\printfnsymbol{1} \and
Songyang Zhang\textsuperscript{\rm 1}\and 
Xuming He\textsuperscript{\rm 1,2}}
\institute{\textsuperscript{\rm 1}School of Information Science and Technology, ShanghaiTech University\\
\textsuperscript{\rm 2}Shanghai Engineering Research Center of Intelligent Vision and Imaging\\
\email{\{liuyf3, zhangxy9, zhangsy1, hexm\}@shanghaitech.edu.cn}}
\authorrunning{Yongfei Liu, Xiangyi Zhang et al.}
\maketitle

In this material, we firstly report the binary-IoU for a clear comparison with previous works in PASCAL-$5^i$ 1-way setting, and then detail the model complexity of our methods.
We futher demonstrate the versatility of our model on multi-way setting, and effectiveness of graph attention network for utilizing unlabeled data, as a supplement to Sec.5.3. Finally, we provide more qualitative visualization results for PASCAL-$5^i$ and COCO-$20^i$.  
\section{Binary-IoU for PASCAL-$5^i$} 
As in Tab.\ref{w1-pas-bin}, our model achieves \textbf{70.90}\%(1-shot) and \textbf{77.45}\%(5-shot) with gain of \textbf{1.0}\% and \textbf{6.95}\% respectively in terms of binary-IoU, compared with PGNet, which validates the superiority of our method when distinguishing complex background. 
\begin{table*}[ht]
  \centering
  \label{w1-pas-bin}
  \caption{{\small Averaged binary-IoU over 4 folds of 1-way setting on PASCAL-$5^i$.}}
  \resizebox{1\textwidth}{!}{
      \begin{tabular}{c|cccccccccccc}
          \hline
          \hline
          Methods&MetaSeg[30] & OSLSM[3] & co-FCN[23] & A-MCG[13]
          & PL[7] & AMP[28] & SG-one[38] & PANet[34] & CANet[37] 
          & PGNet[36] & \textbf{PPNet(RN-50)} & \textbf{PPNet(RN-101)} \\
          \hline
          \hline
          1-shot &-     &61.30 &60.10 &61.20 &61.20 &62.20 &63.90 &66.50 &66.20 &69.90 &69.19 & \textbf{70.90} \\
          5-shot &59.50 &61.50 &60.20 &62.20 &62.30 &63.80 &65.90 &70.70 &69.60 &70.50 &75.76 & \textbf{77.45}\\
          \hline
          \hline 
    \end{tabular}}
\end{table*}

\section{Model Complexity of 1-way 1-shot Setting}
We compare the model complexity with PANet* in 1-way 1-shot setting. We note that the PANet* does not utilize unlabeled data and it is difficult to make direct 
comparison. Instead, we decompose the computation cost into two parts: prototype generation cost, $C_g$ and the inference cost on a query image, $C_q$. Below we report model cost (GFLOPs) 
of the experiment in Tab. \ref{GFLOPs}.

\begin{table*}[ht]
    \center
    \caption{GFLOPs results}
    \label{GFLOPs}
    \resizebox{0.6\textwidth}{!}{
    \begin{tabular}{c|c|c|c}
    \hline
    \hline
    Method&$C_g$&$C_q$&Average of $N$ queries
    \\
    \hline
    \hline
    PANet* &68.47& 68.49& 68.47/$N$+68.49 \\
    Ours (w/o unlabeled)& 69.39& 68.57& 69.39/$N$+68.57 \\
    Ours  & 479.39& 68.57 & 479.39/$N$+68.57\\
    \hline
    \hline
    \end{tabular}}
\end{table*}

While our method uses more FLOPs in prototype generation due to extra unlabeled data, our inference cost is similar to the PANet*. In each task, as the prototype generation needs to be computed only once, the inference cost will dominate the average computation cost for sufficient number of queries.

\section{More Quantitative Results on COCO-$20^i$}
\subsection{Evaluation on Multi-Way Setting}
To demonstrate the model's versatility, we perform more experiments of 2-way and 5-way setting on COCO-$20^i$. As shown in Tab.\ref{w1-coco}, the results show that our model outperforms the baseline model PANet* by a sizeable margin both with/without unlabeled data. Even in the more challenging 5-way 1-shot setting, our model still improves mean-IoU consistently in each fold.

\begin{table*}[ht]
    \center
    \caption{{\small Mean-IoU results of \textbf{2-way 1-shot} and \textbf{5-way 1-shot} on COCO-$20^i$ \textit{split}-A. Red numbers denote the averaged mean-IoU over 4 folds.}}
    \label{w1-coco}
    \resizebox{0.8\textwidth}{!}{
    \begin{tabular}{c|c|ccccc|ccccc}
    \hline
    \hline
    \multirow{2}{*}{Methods}&\multirow{2}{*}{Backbone}&\multicolumn{5}{c|}{\textbf{2-way, 1-shot}}&\multicolumn{5}{c}{\textbf{5-way, 1-shot}}
    \\
    &&fold-1&fold-2 &fold-3&fold-4 &{\color[rgb]{0.7725,0.3529,0.06667}mean}&fold-1&fold-2 &fold-3&fold-4 &{\color[rgb]{0.7725,0.3529,0.06667}mean}\\
    \hline
    \hline
    PANet~[34]                                &VGG16   &29.88&21.13&20.46&15.37&{\color[rgb]{0.7725,0.3529,0.06667}21.71}  &24.94&19.85&19.28&14.11&{\color[rgb]{0.7725,0.3529,0.06667}19.55} \\
    PANet*~[34]                               &RN50    &31.86&21.47&21.31&16.43&{\color[rgb]{0.7725,0.3529,0.06667}22.76}  &27.20&21.50&19.66&15.35&{\color[rgb]{0.7725,0.3529,0.06667}20.93}\\
    PPNet(\textbf{w/o $\mathcal{S}^u$})       &RN50    &33.87&23.98&22.75&17.59&{\color[rgb]{0.7725,0.3529,0.06667}24.55}  &29.12&22.29&21.10&16.37&{\color[rgb]{0.7725,0.3529,0.06667}22.22}\\
    PPNet                                     &RN50    &\textbf{34.20}&\textbf{24.21}&\textbf{23.39}&\textbf{19.06}&{\color[rgb]{0.7725,0.3529,0.06667}\textbf{25.22}}  &\textbf{30.84}&\textbf{23.03}&\textbf{21.32}&\textbf{17.93}&{\color[rgb]{0.7725,0.3529,0.06667}\textbf{23.28}}\\
    \hline
    \hline
    \end{tabular}}
    \label{w1-coco}
\end{table*}

\subsection{More Investigation of Graph Attention Network}
We compare the effectiveness of the graph attention network in Sec.4.2 with a non-parametric graph attention network for utilizing unlabeled data. Here the non-parametric graph attention network means the network takes cosine distance as similarity function $d$ in Equ.4, 5, and removes the linear mapping weight $\mathbf{W}$ in Equ.4. As in Tab.\ref{aba}, the performance will drop from 27.16\% to 26.32\%, which suggests that our graph attention network encode meta-knowledge of message propagation, and are effective for capturing informative features from unlabeled data.

\begin{table*}[ht]
    \center
    \caption{{\small Ablation studies of 1-way 1-shot on COCO-$20^i$ split-A in every fold. Red numbers denote the averaged mean-IoU over 4 folds.}}
    \label{aba}
    \resizebox{0.75\textwidth}{!}{
        \begin{tabular}{c|ccc|ccccc}
            \hline
            \hline
            &&&&\multicolumn{5}{c}{1-shot}
            \\
            Model                                &PAP&SEM&UD &fold-1&fold-2 &fold-3&fold-4 &{\color[rgb]{0.7725,0.3529,0.06667}mean}\\
            \hline
            \hline
            Baseline (PANet*)                    &-&-&-&31.50&22.58&21.50&16.20&{\color[rgb]{0.7725,0.3529,0.06667}22.95} \\
                                                 &\checkmark&\checkmark&\checkmark(\color{blue}{w/o params}) &36.34&23.59&26.39&18.97&{\color[rgb]{0.7725,0.3529,0.06667}26.32} \\
            PPNet                                 &\checkmark&\checkmark&\checkmark&\textbf{36.48}&\textbf{26.53}&\textbf{25.99}&\textbf{19.65}&{\color[rgb]{0.7725,0.3529,0.06667}\textbf{27.16}} \\
            \hline
            \hline
            \end{tabular}
        
    }
\end{table*}

\section{More Qualitative Visualization for PASCAL-$5^i$ and COCO-$20^i$}
\subsection{Visualization for PASCAL-$5^i$}
\begin{figure}
    \centering
    \includegraphics[width=1\linewidth]{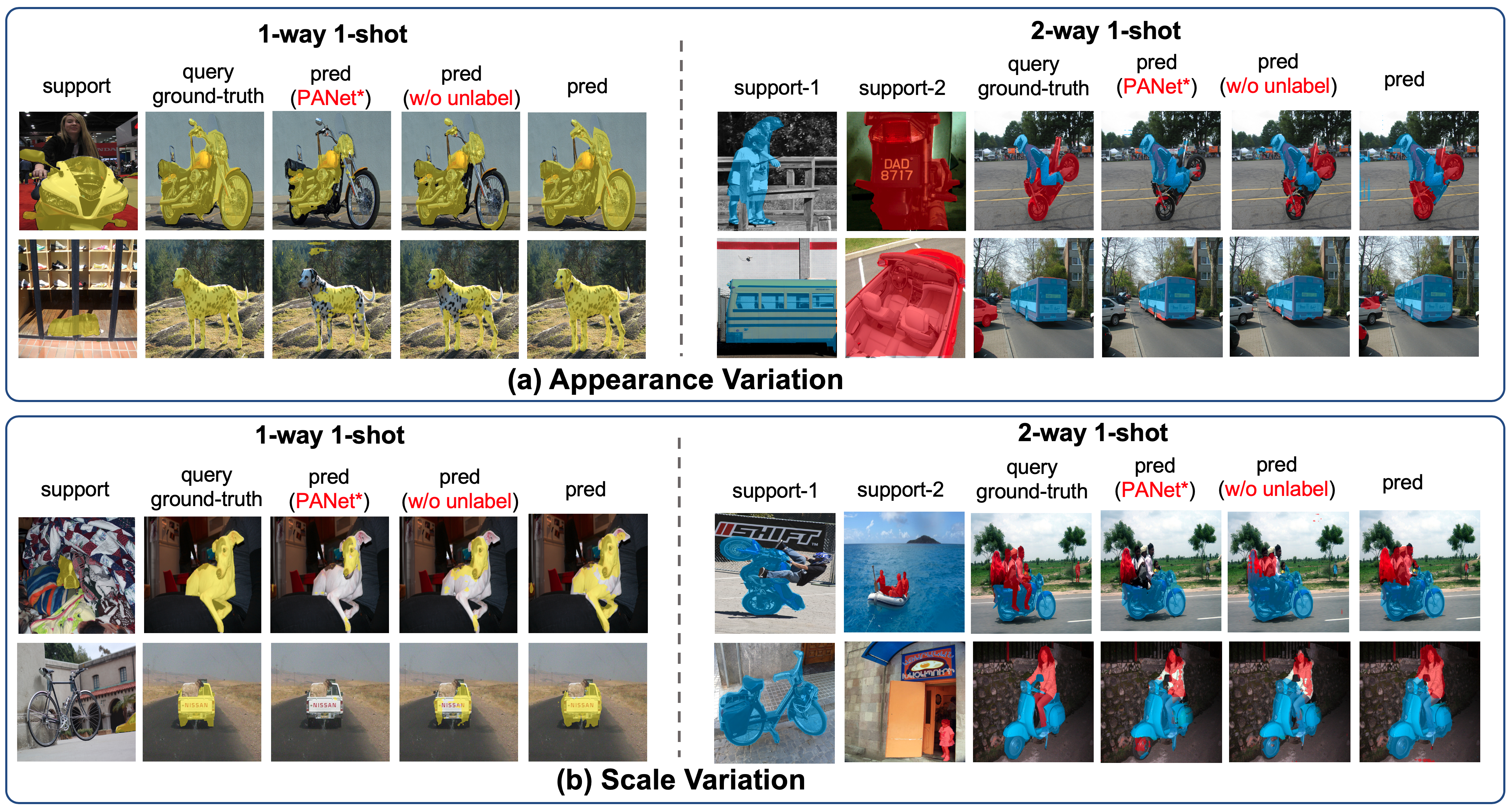}
    \caption{{\small Qualitative Visualization of 1-way 1-shot and  2-way 1-shot on PASCAL-$5^i$. (a) demonstrates the prediction results in the appearance variation scenario while (b) shows the prediction in scale variation}}
    \label{demo-pas}
\end{figure}

As in Fig.\ref{demo-pas}, our model can cope with the large appearance and scale variations between support and query images by utilizing the unlabeled data, both in 1-way 1-shot and 2-way 1-shot setting.

\subsection{Visualization for COCO-$20^i$}
\begin{figure}
    \centering
    \includegraphics[width=\linewidth]{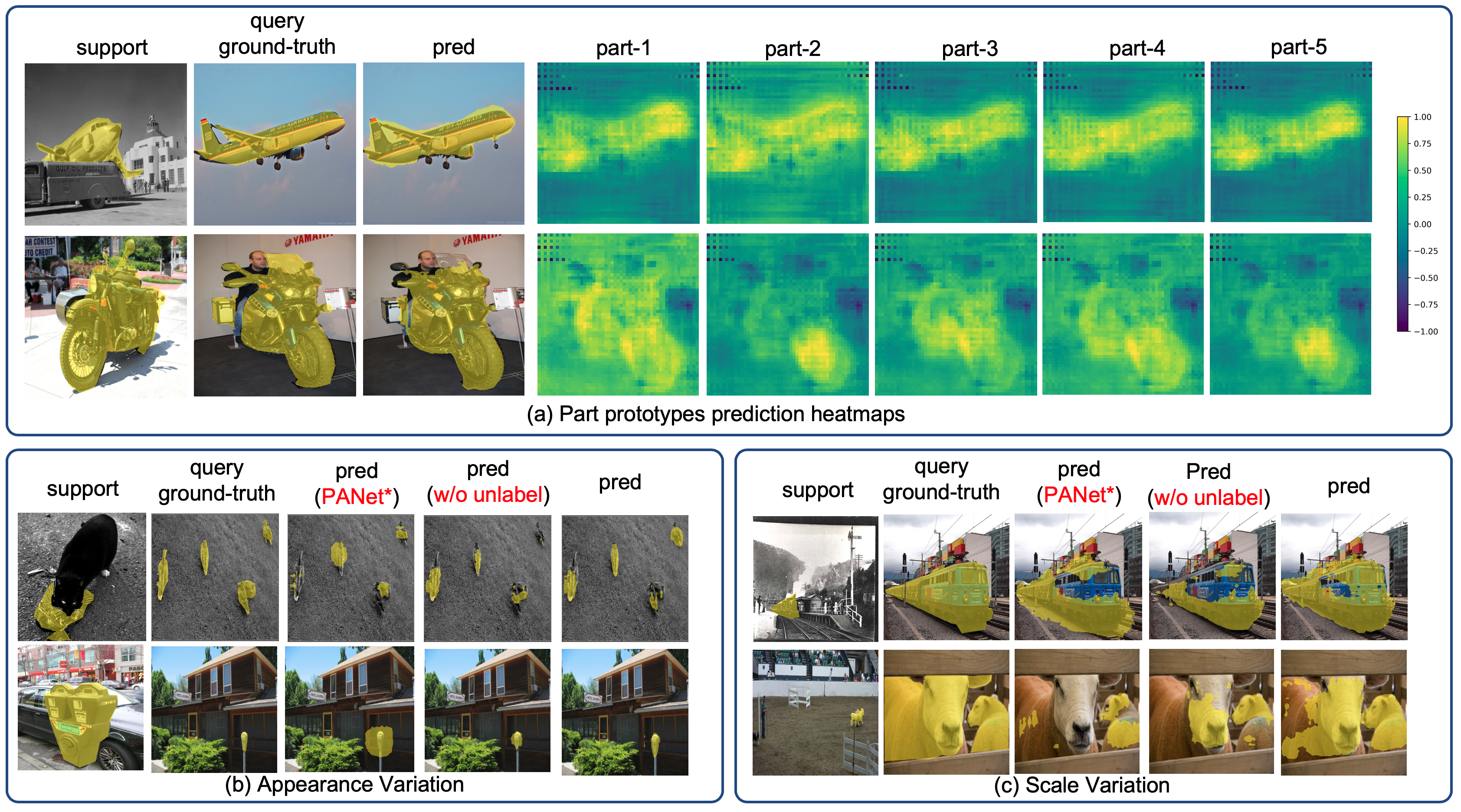}
    \caption{{\small Qualitative Visualization of 1-way 1-shot on COCO-$20^i$ \textit{split}-A. (a) shows the part prototypes prediction heatmaps. The prediction results in the appearance variation and scale variation scenario are demonstrated in (b),(c) respectively.}}
    \label{demo-coco}
\end{figure}

We also provide more visualization results of 1-way 1-shot setting for COCO-$20^i$ as in Fig.\ref{demo-coco}. Our part-aware prototype network is still capable of modeling one semantic class at a fine-grained level and further coping with variations between support and query images in this more challenging benchmark.

\clearpage
%
%